\pdfoutput=1
\documentclass[11pt]{article}

\usepackage[]{naacl2021}

\usepackage{times}
\usepackage{latexsym}
\usepackage{multirow}
\usepackage{float}
\usepackage{booktabs}
\usepackage{graphicx}
\usepackage{pgfplots}
\usepackage{enumitem}

\pgfplotsset{
tick label style={font=\small},
label style={font=\small},
title style={font=\normalsize},
legend style={font=\footnotesize}
}
\newcommand\scroll{\includegraphics[height=0.75em]{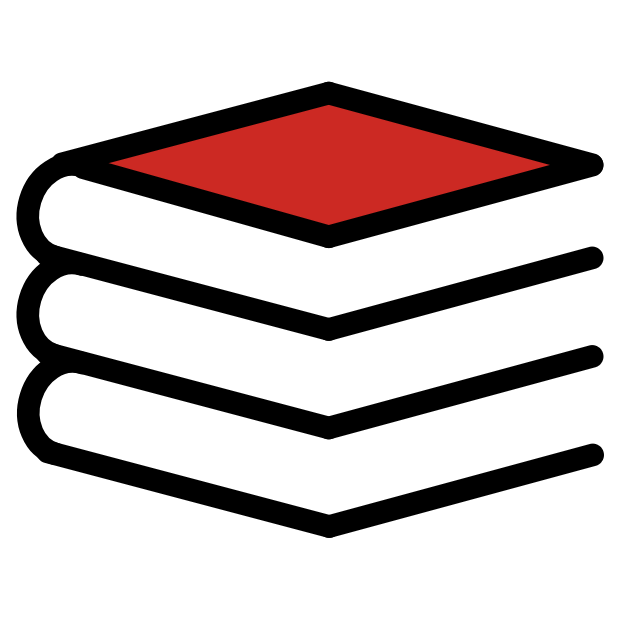}}
\newcommand\baby{\includegraphics[height=0.75em]{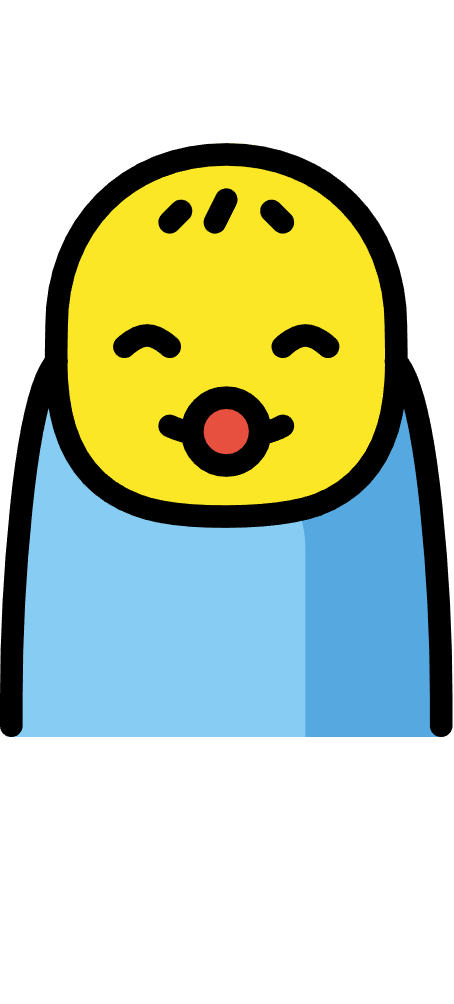}\includegraphics[height=0.75em]{fables/baby.png}\includegraphics[height=0.75em]{fables/baby.png}}
\newcommand\petri{\includegraphics[height=0.75em]{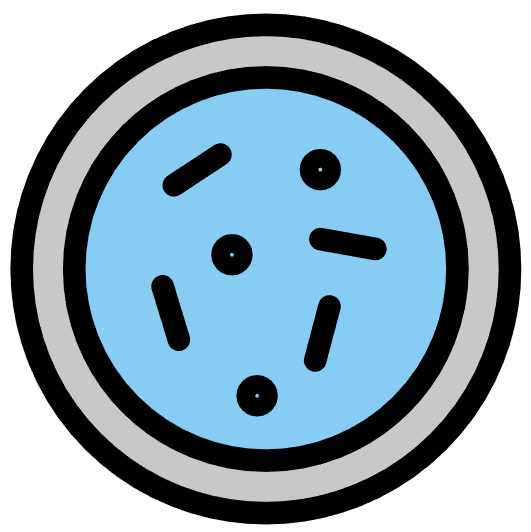}}
\newcommand\curriculum{\includegraphics[height=0.6em]{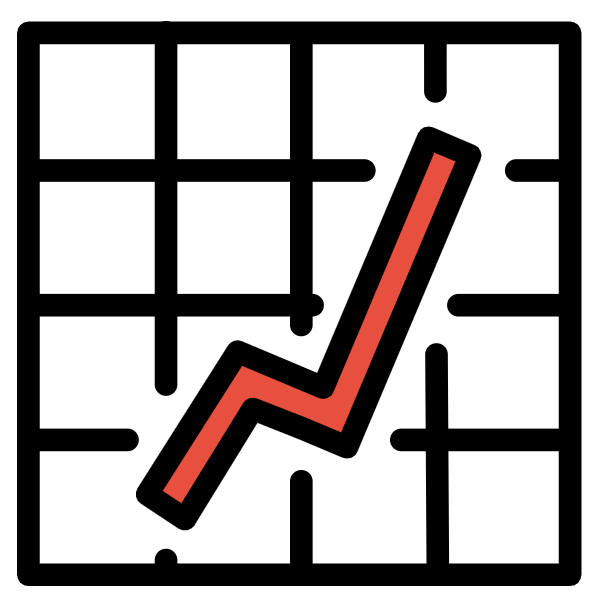}}

\usepackage[T1]{fontenc}
\usepackage[utf8]{inputenc}

\usepackage{microtype}

\title{Few-Shot Text Classification with Triplet Networks, \\ Data Augmentation, and Curriculum Learning}

\author{
Jason Wei$^{\scroll}$ \hspace{1.3mm} 
Chengyu Huang$^{\baby}$ \hspace{1.3mm} 
Soroush Vosoughi$^{\petri}$ \hspace{1.3mm} 
Yu Cheng$^{\curriculum}$ \hspace{1.3mm} 
Shiqi Xu$^{\scroll}$ \\
  \scroll ProtagoLabs \hspace{0.3cm} \baby International Monetary Fund \\
  \petri Dartmouth College \hspace{0.3cm} \curriculum Microsoft AI \\
  \texttt{\{jason,xu\}@protagolabs.com} \hspace{0.3cm} \texttt{huangchengyu24@gmail.com} \\
  \texttt{soroush@dartmouth.edu} \hspace{0.3cm} \texttt{yu.cheng@microsoft.com}\\
  } 

\date{}

\def\shortequals{\hspace{0.4mm}$=$\hspace{0.4mm}}
\def\shortpm{\hspace{0.4mm}$\pm$\hspace{0.4mm}}

\begin{document}
\maketitle
\begin{abstract}
Few-shot text classification is a fundamental NLP task in which a model aims to classify text into a large number of categories, given only a few training examples per category.

This paper explores data augmentation---a technique particularly suitable for training with limited data---for this few-shot, highly-multiclass text classification setting.
On four diverse text classification tasks, we find that common data augmentation techniques can improve the performance of triplet networks by up to 3.0\% on average.

To further boost performance, we present a simple training strategy called \textit{curriculum data augmentation}, which leverages curriculum learning by first training on only original examples and then introducing augmented data as training progresses.
We explore a \textit{two-stage} and a \textit{gradual} schedule, and find that, compared with standard single-stage training, curriculum data augmentation trains faster, improves performance, and remains robust to high amounts of noising from augmentation.
\end{abstract}

\section{Introduction}

Traditional text classification tasks such as sentiment classification \cite{socher-etal-2013-recursive} typically have few output classes (e.g., in binary classification), each with many training examples.
Many practical scenarios such as relation classification \cite{han-etal-2018-fewrel}, answer selection \cite{kumar-etal-2019-improving}, and sentence clustering \cite{mnasri-etal-2017-taking}, however, have a converse setup characterized by a large number of output classes \cite{10.5555/2627435.2638582}, often with few training examples per class. This scenario, which we henceforth refer to as \textit{few-shot, highly-multiclass text classification}, is a common setting in NLP applications and can be challenging due to the scarcity of training data.
\begin{figure}[ht]
    \centering
    \includegraphics[width=\linewidth]{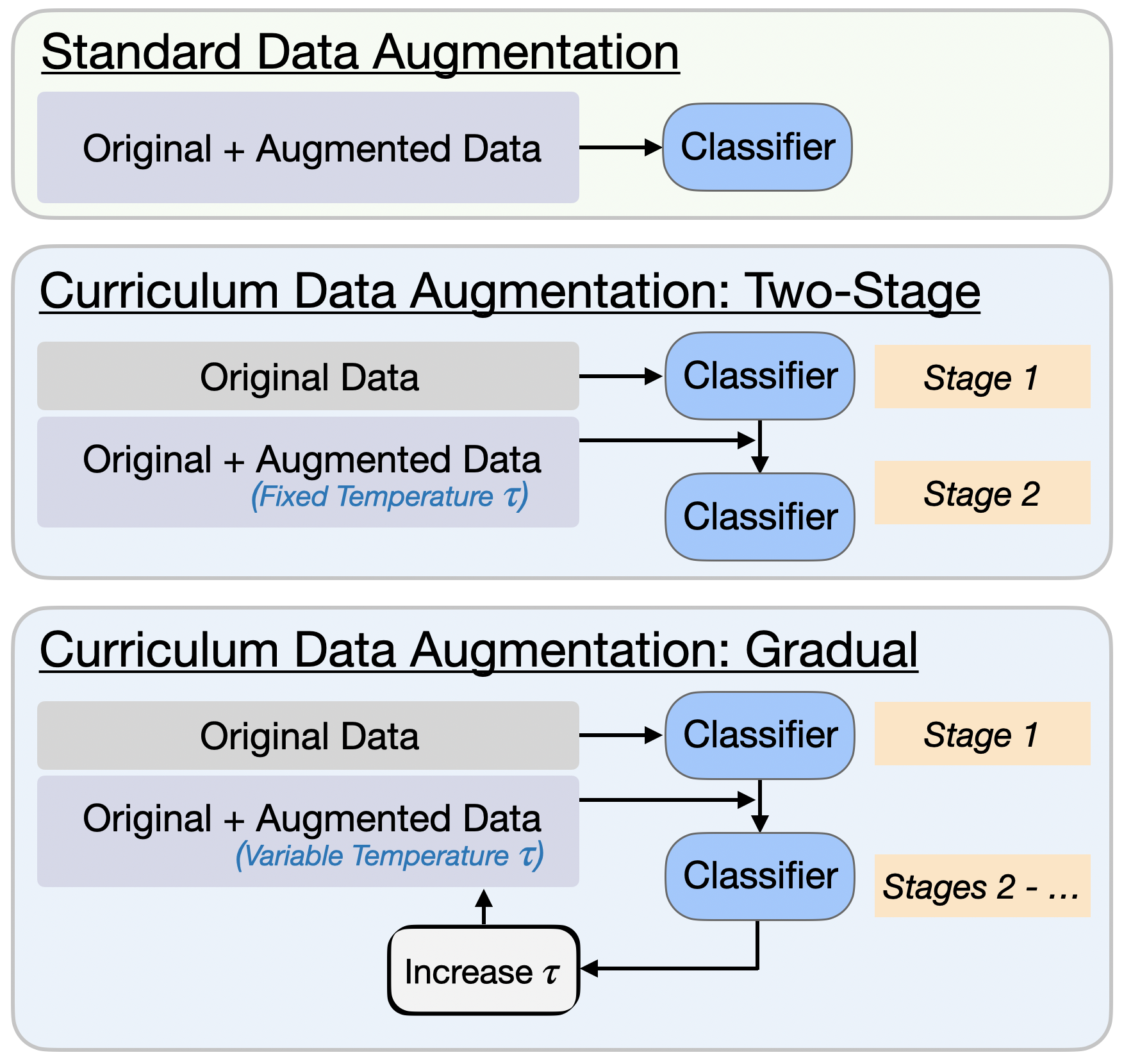}
    \vspace{-5.3mm}
    \caption{Schematic showing the two types of curriculum augmentation that we propose. 
    $\tau$ is a parameter that controls augmentation temperature (fraction of perturbed tokens).}
    \vspace{-1mm}
    \label{fig:pull}
\end{figure}

Data augmentation for NLP has seen increased interest in recent years \cite{wei-zou-2019-eda,10.1145/3366424.3383552}. %,lewis-etal-2020-bart
In traditional text classification tasks, it has been shown that although performance improvements can be marginal when training data is sufficient, augmentation is especially beneficial in limited data scenarios \cite{xie2020unsupervised}. %,andreas-2020-good}.
As such, we hypothesize that the few-shot, highly-multiclass text classification scenario is a suitable context for data augmentation.

Based on this motivation, our paper makes two main contributions.
\begin{itemize}[leftmargin=*]
    \itemsep0em
    \item First, we apply popular data augmentation techniques to the common triplet loss \cite{facenet} approach for few-shot, highly multiclass classification, finding that out-of-the-box augmentation can improve performance noticeably.
    \item We then propose a simple curriculum learning strategy called \textit{curriculum data augmentation} and experiment with two schedules, as shown in Figure \ref{fig:pull}.
    A \textit{two-stage} curriculum, which first trains on original data and then introduces augmented data of fixed temperature (amount of noising), achieves slightly better performance than standard augmentation, while training faster and remaining more robust to high temperatures.
    A \textit{gradual} curriculum, which also first trains on original data only but gradually increases augmentation temperature at each subsequent stage, takes longer to converge but improves more than 1\% over standard augmentation.
\end{itemize}

\section{Curriculum Data Augmentation}

\noindent \textbf{Motivation.} Inspired by human and animal learning, curriculum learning \cite{Bengio2009} posits that neural networks train better when examples are not randomly presented but instead organized in a meaningful order that gradually shows more concepts and complexity. 
Traditionally, curriculum learning approaches first assume that a range of example difficulty exists in the data and then leverage various heuristics to sort examples by difficulty and train models on progressively harder examples \cite{Bengio2009,Tsvetkov2016,Weinshall2018}.
A newer school of thought, however, has noted that instead of discovering a curriculum in existing data, data can be intentionally modified to dictate an artificial range of difficulty \cite{Korbar2018,ganesh2020rethinking}---
this is the approach we will take here.

\vspace{1.3mm} \noindent \textbf{Our approach.}
Unlike data augmentation in computer vision where augmented data undoubtedly resembles original data, in text, data augmentation techniques might introduce linguistic adversity and therefore can be seen as a form of noising \cite{li-etal-2017-robust,wang-etal-2018-switchout}, where noised data is harder to learn from than unmodified original data.
As such, we can create an artificial curriculum in the data by leveraging controlled application of data augmentation, starting by training on only original data and then adding augmented data with a higher levels of noising as training progresses. 
Specifically, we propose two simple schedules. \textbf{(1) Two-stage} curriculum data augmentation calls for one stage of training with only original data, followed by one stage of training with augmented data of fixed temperature. \textbf{(2) Gradual} curriculum data augmentation involves one stage of training with only original data, followed by multiple stages of training with augmented data where the temperature of augmented data (i.e., fraction of perturbed tokens) gradually increases each stage. 
\pgfplotsset{width=6.5cm,height=5.4cm,compat=1.9}
\begin{figure}[ht]
\begin{centering}
\begin{tikzpicture}
\begin{axis}[
    title style={yshift=-1.5ex,},
    xlabel={Updates (thousands)},
    ylabel={Validation Accuracy (\%)},
    xmin=0, xmax=24.4,
    ymin=8.6, ymax=20.3,
    xtick={0, 4, 8, 12, 16, 20, 24},
    ytick={10, 12, 14, 16, 18, 20},
    legend pos=south east,
    ymajorgrids=true,
    xmajorgrids=true,
    grid style=dashed,
    x label style={at={(axis description cs:0.5,-0.1)},anchor=north},
    y label style={at={(axis description cs:-0.1,0.5)},anchor=south},
]
\addplot[
    color=violet,
    mark=o,
    mark size=0.6pt,
    ]
    coordinates {
    (0,  6.92)
    (0.3,  11.78)
    (0.6,  13.69)
    (0.9,  14.78)
    (1.2,  15.45)
    (1.5,  15.82)
    (1.8,  15.88)
    (2.1,  16.22)
    (2.4,  16.38)
    (2.7,  16.52)
    (3,  16.71)
    (3.3,  16.8)
    (3.6,  16.95)
    (3.9,  16.94)
    (4.2,  17.22)
    (4.5,  17.49)
    (4.8,  17.51)
    (5.1,  17.62)
    (5.4,  17.7)
    (5.7,  17.88)
    (6,  17.9)
    (6.3,  17.9)
    (6.6,  17.86)
    (6.9,  17.92)
    (7.2,  17.92)
    (7.5,  17.83)
    (7.8,  17.83)
    (8.1,  17.9)
    (8.4,  18.15)
    (8.7,  18.3)
    (9,  18.35)
    (9.3,  18.4)
    (9.6,  18.37)
    (9.9,  18.42)
    (10.2,  18.51)
    (10.5,  18.24)
    (10.8,  18.27)
    (11.1,  18.3)
    (11.4,  18.38)
    (11.7,  18.27)
    (12,  18.21)
    (12.3,  18.72)
    (12.6,  18.92)
    (12.9,  18.79)
    (13.2,  18.92)
    (13.5,  18.79)
    (13.8,  18.51)
    (14.1,  18.59)
    (14.4,  18.83)
    (14.7,  18.79)
    (15,  18.67)
    (15.3,  18.78)
    (15.6,  18.81)
    (15.9,  18.74)
    (16.2,  18.9)
    (16.5,  18.99)
    (16.8,  18.94)
    (17.1,  18.87)
    (17.4,  18.72)
    (17.7,  18.72)
    (18,  18.8)
    (18.3,  18.75)
    (18.6,  18.63)
    (18.9,  18.65)
    (19.2,  18.74)
    (19.5,  18.6)
    (19.8,  18.52)
    (20.1,  18.84)
    (20.4,  19.12)
    (20.7,  19.02)
    (21,  19.11)
    (21.3,  19.02)
    (21.6,  19.09)
    (21.9,  19.05)
    (22.2,  19.05)
    (22.5,  19.15)
    (22.8,  19.05)
    (23.1,  19.03)
    (23.4,  18.96)
    (23.7,  19.01)
    (24,  18.91)
    (24.3,  18.87)
    (24.6,  18.89)
    (24.9,  18.)
    };
    \addlegendentry{Curriculum: gradual}
\addplot[
    color=blue,
    mark=o,
    mark size=0.6pt,
    ]
    coordinates {
    (0,  7.12)
    (0.3,  11.27)
    (0.6,  13.38)
    (0.9,  14.82)
    (1.2,  14.98)
    (1.5,  15.78)
    (1.8,  16)
    (2.1,  16.32)
    (2.4,  16.26)
    (2.7,  16.32)
    (3,  16.69)
    (3.3,  16.42)
    (3.6,  16.65)
    (3.9,  16.65)
    (4.2,  17.01)
    (4.5,  17.26)
    (4.8,  17.49)
    (5.1,  17.7)
    (5.4,  17.71)
    (5.7,  17.8)
    (6,  17.87)
    (6.3,  17.88)
    (6.6,  18.03)
    (6.9,  17.75)
    (7.2,  17.95)
    (7.5,  17.88)
    (7.8,  17.92)
    (8.1,  18)
    (8.4,  17.97)
    (8.7,  17.92)
    (9,  17.87)
    (9.3,  18.09)
    (9.6,  18.02)
    (9.9,  17.84)
    (10.2,  17.92)
    (10.5,  17.99)
    (10.8,  18)
    (11.1,  18.03)
    (11.4,  18.15)
    (11.7,  18.13)
    (12,  18.05)
    (12.3,  18.01)
    (12.6,  18.06)
    (12.9,  18.07)
    (13.2,  17.99)
    (13.5,  17.92)
    (13.8,  17.98)
    (14.1,  17.99)
    (14.4,  17.92)
    (14.7,  17.92)
    (15,  18.02)
    (15.3,  18.02)
    (15.6,  18.06)
    (15.9,  17.91)
    (16.2,  17.9)
    (16.5,  17.74)
    (16.8,  17.88)
    (17.1,  17.84)
    (17.4,  17.77)
    (17.7,  17.68)
    (18,  17.8)
    (18.3,  17.8)
    (18.6,  17.7)
    (18.9,  17.67)
    (19.2,  17.68)
    (19.5,  17.81)
    (19.8,  17.62)
    (20.1,  17.59)
    (20.4,  17.42)
    (20.7,  17.49)
    (21,  17.39)
    (21.3,  17.62)
    (21.6,  17.52)
    (21.9,  17.52)
    (22.2,  17.59)
    (22.5,  17.49)
    (22.8,  17.28)
    (23.1,  17.52)
    (23.4,  17.47)
    (23.7,  17.56)
    (24,  17.47)
    (24.3,  17.35)
    (24.6,  17.45)
    (24.9,  17.3)
    };
    \addlegendentry{Curriculum: two-stage}
\addplot[
    color=red,
    mark=o,
    mark size=0.6pt,
    ]
    coordinates {
    (0,  7.04)
    (0.3,  10.42)
    (0.6,  12.25)
    (0.9,  13.82)
    (1.2,  14.7)
    (1.5,  15.04)
    (1.8,  15.35)
    (2.1,  15.98)
    (2.4,  15.92)
    (2.7,  15.93)
    (3,  16.05)
    (3.3,  16.26)
    (3.6,  16.2)
    (3.9,  16.38)
    (4.2,  16.48)
    (4.5,  16.46)
    (4.8,  16.65)
    (5.1,  16.78)
    (5.4,  16.83)
    (5.7,  16.83)
    (6,  16.88)
    (6.3,  16.95)
    (6.6,  17.12)
    (6.9,  17.16)
    (7.2,  17.15)
    (7.5,  17.41)
    (7.8,  17.54)
    (8.1,  17.38)
    (8.4,  17.45)
    (8.7,  17.29)
    (9,  17.36)
    (9.3,  17.45)
    (9.6,  17.57)
    (9.9,  17.72)
    (10.2,  17.6)
    (10.5,  17.65)
    (10.8,  17.77)
    (11.1,  17.78)
    (11.4,  17.65)
    (11.7,  17.84)
    (12,  17.91)
    (12.3,  17.68)
    (12.6,  17.84)
    (12.9,  17.72)
    (13.2,  17.85)
    (13.5,  17.74)
    (13.8,  17.78)
    (14.1,  17.78)
    (14.4,  17.75)
    (14.7,  17.7)
    (15,  17.62)
    (15.3,  17.67)
    (15.6,  17.6)
    (15.9,  17.73)
    (16.2,  17.72)
    (16.5,  17.52)
    (16.8,  17.57)
    (17.1,  17.62)
    (17.4,  17.65)
    (17.7,  17.58)
    (18,  17.73)
    (18.3,  17.53)
    (18.6,  17.39)
    (18.9,  17.45)
    (19.2,  17.44)
    (19.5,  17.43)
    (19.8,  17.5)
    (20.1,  17.38)
    (20.4,  17.45)
    (20.7,  17.45)
    (21,  17.42)
    (21.3,  17.13)
    (21.6,  17.33)
    (21.9,  17.49)
    (22.2,  17.45)
    (22.5,  17.55)
    (22.8,  17.48)
    (23.1,  17.39)
    (23.4,  17.28)
    (23.7,  17.08)
    (24,  17.22)
    (24.3,  17.22)
    (24.6,  17.34)
    (24.9,  17.)
    };
    \addlegendentry{Standard Augmentation}
\addplot[
    color=black,
    mark=o,
    mark size=0pt,
    ]
    coordinates {
    (0,  6.92)
    (0.3,  11.78)
    (0.6,  13.69)
    (0.9,  14.78)
    (1.2,  15.45)
    (1.5,  15.82)
    (1.8,  15.88)
    (2.1,  16.22)
    (2.4,  16.38)
    (2.7,  16.52)
    (3,  16.71)
    (3.3,  16.8)
    (3.6,  16.95)
    (3.9,  16.94)
    (4.2,  16.95)
    (4.5,  16.96)
    (4.8,  17.06)
    (5.1,  17.09)
    (5.4,  17.1)
    (5.7,  16.98)
    (6,  17.1)
    (6.3,  17.01)
    (6.6,  16.85)
    (6.9,  16.82)
    (7.2,  16.52)
    (7.5,  16.8)
    (7.8,  16.67)
    (8.1,  16.74)
    (8.4,  16.81)
    (8.7,  16.82)
    (9,  16.57)
    (9.3,  16.45)
    (9.6,  16.65)
    (9.9,  16.58)
    (10.2,  16.46)
    (10.5,  16.5)
    (10.8,  16.44)
    (11.1,  16.38)
    (11.4,  16.27)
    (11.7,  16.48)
    (12,  16.42)
    (12.3,  16.05)
    (12.6,  16.06)
    (12.9,  16.08)
    (13.2,  16.2)
    (13.5,  16.16)
    (13.8,  16.21)
    (14.1,  16.17)
    (14.4,  16.25)
    (14.7,  16.15)
    (15,  16.14)
    (15.3,  16.29)
    (15.6,  16.08)
    (15.9,  15.94)
    (16.2,  16.19)
    (16.5,  16.14)
    (16.8,  16.07)
    (17.1,  16.12)
    (17.4,  15.99)
    (17.7,  16.29)
    (18,  15.98)
    (18.3,  15.95)
    (18.6,  15.91)
    (18.9,  15.82)
    (19.2,  16.16)
    (19.5,  16.05)
    (19.8,  16.06)
    (20.1,  15.89)
    (20.4,  15.95)
    (20.7,  16.11)
    (21,  16.01)
    (21.3,  16.11)
    (21.6,  16.1)
    (21.9,  15.97)
    (22.2,  15.97)
    (22.5,  15.95)
    (22.8,  15.99)
    (23.1,  16.08)
    (23.4,  16)
    (23.7,  15.68)
    (24,  15.98)
    (24.3,  15.98)
    (24.6,  15.97)
    (24.9,  15.9)
    };
    \addlegendentry{No Augmentation}
\end{axis}
\end{tikzpicture}
% \vspace{-2.5mm}
\caption{
Example training plot for the HUFF dataset (41 classes, with 5 examples per class) using EDA augmentation \cite{wei-zou-2019-eda}.
Our proposed two-stage curriculum (second stage starts at four-thousand updates) trains faster and achieves slightly higher performance compared with standard augmentation while using the same number of updates.
Our proposed gradual curriculum (which here linearly increases augmentation temperature $\tau$ by 0.1 at $\{4, 8, 12, 16, 20\}$-thousand updates) outperforms both standard augmentation and the two-stage curriculum, but takes longer to converge. 
Results shown are averaged over thirteen random seeds. 
}
% \vspace{-4.5mm}
\label{fig:training}
\end{centering}
\end{figure}
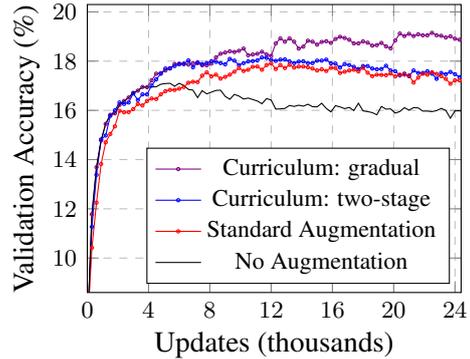

\section{Experimental Setup}

We conduct empirical experiments to evaluate curriculum data augmentation on a variety of text classification tasks using a triplet loss model.\footnote{Code is made publicly available at \url{https://github.com/jasonwei20/triplet-loss}.}

\subsection{Datasets} 
We will consider four diverse few-shot, highly-multiclass text classification scenarios:
\begin{enumerate}[leftmargin=*]
    \itemsep0em
    \item \textbf{HUFF} ($c=41$). The HuffPost Dataset categorizes 200k news headlines from 2012--2018 into 41 categories such as politics, wellness, entertainment, and travel \cite{huff}. We use all 41 categories and perform a 70\%-30\% train-test split by class.
    \item \textbf{FEWREL} ($c=64$). The FewRel dataset contains sentences categorized by a relationship between its specified head and tail tokens such as `capital of,' `member of,' and `birth name' \cite{han-etal-2018-fewrel}. We use all 64 classes given in the posted training set, splitting 100 examples per class into a test set, with the remainder of the examples going into the training set.
    \item \textbf{COV-C} ($c=87$). The COVID-Q dataset classifies questions into 89 clusters where all questions in a cluster ask about the same thing  \cite{wei2020people}. We use the train-test split with three training examples per class as given by the authors. We find that 2 of the 89 classes in the training set actually have only two examples per class instead of the reported three, and so we remove these classes from the training and test sets and use the 87 classes that remain.
    \item \textbf{AMZN} ($c=318$). The Amazon product review dataset aims to categorize a product into a certain class given a review \cite{amzn}. We only consider the 318 `level-3' classes given in this dataset with at least six examples per class.
\end{enumerate}
To balance the class distribution during experiments, we randomly sample $N_c$ examples per class to be used for training, with $N_c$ varying based on the experiment and dataset. 
Our sampled training sets for COV-C and AMZN have $N_c$\shortequals3 examples per class, and our training sets for HUFF and FEWREL have $N_c$\shortequals10, a common low-resource scenario.\footnote{For COV-C, the given training set size is $N_c$\shortequals3, and for AMZN, $N_c$\shortequals3 is the largest possible such that the training set is balanced by class.} For all experiments, we use top-1 accuracy (\%) as the evaluation metric.

\subsection{Triplet Loss Model}
For few-shot, highly-multiclass classification, a common approach is the \textit{triplet loss} classifier \cite{facenet}, first developed for facial recognition and now also used in NLP \cite{DBLP:journals/corr/SantosTXZ16,ein-dor-etal-2018-learning,lauriola2020context}.
Specifically addressing few-shot classification, a triplet loss network minimizes distance between examples with the same label and maximizes distance between examples with different labels. 
During training, given a triplet of (anchor $a$, positive example $p$, and negative example $n$), a triplet loss network minimizes:
\begin{equation}
    L = \sum_{i} d(a, p) - d(a, n) + \alpha \ , 
\end{equation}
where $\alpha$ is a margin enforced between positive and negative pairs, and $d(\cdot)$ computes the distance between the input encodings of two examples.
To sample triplets, we will consider two strategies: \textit{random sampling}, which selects triplets randomly, and \textit{hard negative mining} \cite{facenet}, where triplets are sampled such that $d(a, p) + \alpha > d(a, n)$.
At evaluation time, a triplet loss classifier returns the class of the example in the training set with the smallest distance to a given test example.
Indeed, both triplet loss and data augmentation target training with limited data, and so combining them seems particularly promising for the the few-shot classification scenario.

For our model, we use standard BERT-base with average-pooled encodings and then train a two-layer triplet loss network on top of these encodings.
Our triplet loss network architecture contains a linear layer with 200 hidden units, tanh activation, a dropout layer with $p=0.4$, and a final linear layer with 40 hidden units.
We use cosine distance, a margin of $\alpha\shortequals0.4$, a batch size of 64 triplets, and a learning rate of $2\times10^{-5}$.

\subsection{Augmentation Techniques}
We implement EDA \cite{wei-zou-2019-eda}, a popular combination of token-level augmentation techniques (synonym replacement, random insertion, random swap, random deletion) that defines their temperature parameter $0 \leq \tau \leq 1$ as the fraction of perturbed tokens, in $\S$\ref{subsec:cda}--\ref{subsec:schedules}, and explore four other techniques in $\S$\ref{subsec:various}.

\subsection{Schedules}
For the two-stage curriculum, we started by training on original data only, and when validation loss converges, we introduce augmented data of fixed temperature at an augmented to original data ratio of 4:1.
For the gradual curriculum, 
we begin with a temperature of $\tau$\shortequals0.0 (equivalent to no augmentation) and then linearly increase the temperature by 0.1 every time validation loss plateaus, up to a final temperature of 0.5. Schedules for each dataset are shown in the Appendix.
Figure \ref{fig:training} shows an example training plot with our proposed curriculum schedules.

\begingroup
\begin{table*}[ht]
    \centering
    \small
    \begin{tabular}{l | c c c c | c c }
        % \toprule
        & HUFF & FEWREL & COV-C & AMZN &  \\
        & $c$\shortequals$41$ & $c$\shortequals$64$ & $c$\shortequals$87$ & $c$\shortequals$318$ & Average & $\Delta$\\
        \midrule
        Cross-entropy loss                                      & 13.3\shortpm2.1 & 32.4\shortpm2.3 & 26.1\shortpm0.8 & 2.0\shortpm0.3 & 18.5 & - \\
        \hspace{2mm} + standard data augmentation               & 16.3\shortpm2.4 & 33.0\shortpm1.1 & 24.0\shortpm1.6 & 2.2\shortpm0.4 & 18.9 & +0.4 \\
        \midrule
        Triplet loss with random sampling                       & 20.9\shortpm1.0 & 43.6\shortpm1.2 & 39.7\shortpm1.0 & 16.4\shortpm1.3 & 30.2 & - \\
        \hspace{2mm} + standard data augmentation               & 22.2\shortpm1.4 & 44.2\shortpm1.6 & 45.4\shortpm1.8 & 16.5\shortpm1.7 & 32.1 & +1.9 \\
        \hspace{2mm} + \textbf{curriculum data augmentation: two-stage}  & 22.3\shortpm1.6 & 44.2\shortpm1.8 & 46.5\shortpm1.7 & 17.2\shortpm1.3 & 32.6 & +2.4\\
        \hspace{2mm} + \textbf{curriculum data augmentation: gradual}    & 23.7\shortpm1.2 & 46.1\shortpm0.9 & 47.1\shortpm1.3 & 17.6\shortpm1.0 & 33.6 & +3.4 \\
        \midrule
        Triplet loss with hard negative mining                  & 21.0\shortpm1.2 & 44.6\shortpm1.2 & 39.5\shortpm1.0 & 16.2\shortpm0.9 & 30.3 & -\\
        \hspace{2mm} + standard data augmentation               & 22.6\shortpm1.8 & 45.0\shortpm1.6 & 48.2\shortpm0.9 & 17.4\shortpm1.7 & 33.3 & +3.0 \\
        \hspace{2mm} + \textbf{curriculum data augmentation: two-stage}  & 22.6\shortpm1.8 & 45.7\shortpm1.4 & 47.6\shortpm1.3 & 17.9\shortpm1.1 & 33.5 & +3.2 \\
        \hspace{2mm} + \textbf{curriculum data augmentation: gradual}    & 23.8\shortpm0.9 & 47.1\shortpm1.4 & 48.9\shortpm0.9 & 18.9\shortpm0.9 & 34.7 & +4.4 \\
        % \bottomrule
    \end{tabular}
    \vspace{-2mm}
    \caption{
    Accuracy (\%) on four diverse highly multiclass classification tasks for no augmentation, standard augmentation, and curriculum augmentation. 
    $c$: number of classes; $\Delta$: improvement compared with no augmentation.
    }
    \label{tab:main_table}
\end{table*}
\endgroup
\section{Results}
\subsection{Curriculum Data Augmentation}
\label{subsec:cda}
This section compares no augmentation, standard augmentation, and curriculum augmentation for triplet loss networks using two different triplet sampling strategies. 
Table \ref{tab:main_table} summarizes these results for five random seeds.
We also implement a cross-entropy loss classifier for reference.

\pgfplotsset{width=5.6cm,height=4.6cm,compat=1.9}
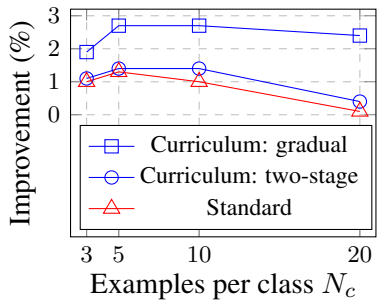
\begin{figure}[th]
\begin{centering}
\begin{tikzpicture}
\begin{axis}[
    xlabel={Examples per class $N_c$},
    ylabel={Improvement (\%)},
    xmin=2, xmax=21,
    ymin=-3.7, ymax=3.2,
    xtick={3, 5, 10, 20},
    ytick={0, 1, 2, 3},
    legend pos=south west,
    ymajorgrids=true,
    xmajorgrids=true,
    grid style=dashed,
    x label style={at={(axis description cs:0.5,-0.115)},anchor=north},
    y label style={at={(axis description cs:-0.08,0.5)},anchor=south},
]
\addplot[
    color=blue,
    mark=square,
    mark size=2.5pt,
    ]
    coordinates {
    (3,     1.9)   
    (5,     2.7)   
    (10,    2.7)   
    (20,    2.4)   
    };
    \addlegendentry{Curriculum: gradual}
\addplot[
    color=blue,
    mark=o,
    mark size=2.5pt,
    ]
    coordinates {
    (3,     1.1)   
    (5,     1.4)   
    (10,    1.4)   
    (20,    0.4)   
    };
    \addlegendentry{Curriculum: two-stage}
\addplot[
    color=red,
    mark=triangle,
    mark size=3.5pt,
    ]
    coordinates {
    (3,     1.0)   
    (5,     1.3)   
    (10,    1.0)   
    (20,    0.1)   
    };
    \addlegendentry{Standard}
    % \node[] at (axis cs: 10,23) {HUFF};
\end{axis}
\end{tikzpicture}
\vspace{-2.5mm}
\caption{
Improvement from data augmentation for different dataset sizes (results averaged over HUFF and FEWREL datasets).
}
\label{fig:dataset_sizes}
\end{centering}
\end{figure}
For triplet loss using random sampling, a model with no augmentation achieved a mean accuracy across our four datasets of 30.2\%, and standard augmentation improved performance noticeably by +1.9\%.
Two-stage curriculum augmentation, which trains for the same number of updates as standard augmentation, achieved a mean accuracy of 32.4\%, outperforming standard augmentation by +0.5\%. 
The gradual curriculum
further improved +1.0\% over the two-stage curriculum.

For triplet loss with hard negative mining, standard augmentation substantially improved +3.0\% over no augmentation, as adding in augmented data, which is more difficult to classify, likely helped generate a more diverse set of hard negatives. 
The two-stage curriculum still maintained small improvement over standard augmentation here, and the gradual curriculum provided an even-stronger boost of +4.4\% over no augmentation, possibly because increasing the temperature of augmented data over time facilitated hard-negative mining more so than using a constant temperature.

Notably, the largest gains for all augmentation types were on COV-C (up to +9.4\%). 
We hypothesize that this occurred not necessarily because of COV-C's smaller data size; rather, there was likely more overfitting to be mitigated by data augmentation as a result of the greater semantic difference between COV-C and the corpus used to pre-train BERT, compared with the other three datasets.

\input{fables/tempature}
\subsection{Ablation: Dataset Size}
\label{sec:ablation-dataset-size}
This ablation investigates how data augmentation performs for different dataset sizes.
Figure \ref{fig:dataset_sizes} shows these results for hard negative mining averaged over HUFF and FEWREL, our two datasets where sufficient data is available. 
The two-stage curriculum outperformed standard augmentation by a small margin, although both dropped in performance at $N_c=20$, consistent with prior findings on the diminished effect of data augmentation for larger datasets \cite{xie2020unsupervised,andreas-2020-good}.
The gradual curriculum, on the other hand, maintained relatively robust improvement for all dataset sizes explored.

\subsection{Ablation: Augmentation Temperature}
\label{subsec:temp}
Effective curriculum learning necessitates a range of difficulty in training data.
In our case, this range is controlled by \textit{augmentation temperature},
a parameter that dictates how perturbed augmented examples are and therefore affects the distribution of difficulty in training examples.
\begingroup
\setlength{\tabcolsep}{3.4pt}
\begin{table}[ht]
    \centering
    \small
    \begin{tabular}{l | c c c c | c}
        Schedule & HUFF & FEWREL & COV-C & AMZN & Avg. \\
        \midrule
        Curriculum & 23.7 & 46.1 & 48.1 & 17.6 & \textbf{33.63} \\
        Control & 23.5 & 45.3 & 46.3 & 17.1 & 33.05 \\
        Anti & 23.3 & 44.8 & 46.2 & 17.5 & 32.95 \\
    \end{tabular}
    \caption{
    Gradual curriculum augmentation with three schedules. 
    Curriculum: temperature $\tau$ increases. 
    Control: $\tau$ is randomly selected every fifty updates. 
    Anti: decreasing $\tau$.
    Results are shown for ten seeds.
    }
    \label{tab:curricula_table}
\end{table}
\endgroup
When the distribution of difficulty in data is larger, we should expect a greater improvement from curriculum learning.

Figure \ref{fig:temperature} compares standard and two-stage curriculum augmentation for various temperatures, with results averaged over all four datasets.
At low temperature, augmented examples remained pretty similar to original examples, and so the range of difficulty in examples was small and therefore curriculum learning showed little improvement.
At higher temperatures, however, augmented examples became quite different from original examples, and so the range of difficulty in examples was much larger and therefore curriculum data augmentation improved over standard augmentation more. 
Whereas \citet{wei-zou-2019-eda} recommend $\tau \in \{0.05, 0.1\}$, our curriculum framework liberates us to use much larger $\tau$ and maintain relatively robust improvements even at $\tau \in \{0.4, 0.5\}$ when standard augmentation is no longer useful.

\subsection{Ablation: Curriculum Schedules}
\label{subsec:schedules}
The gradual curriculum linearly increases temperature $\tau$ from 0.0 to 0.5 in six stages, and so to isolate the effect of this curriculum, in this section we compare it with a control schedule (where the $\tau$ in each stage is decided randomly) and an anti-curriculum schedule (where $\tau$ linearly decreases from 0.5 to 0.0 in six stages). 
As expected, these results, shown in Table \ref{tab:curricula_table}, indicate that the curriculum contributes substantively over the control schedule.

\pgfplotsset{width=6cm, height=7cm}
\begin{figure}[ht]
\begin{tikzpicture}
  \begin{axis}
[
    % title    = Per,
    xbar,
    y axis line style = { opacity = 0 },
    axis x line       = none,
    % x axis line style = { opacity = 0 },
    tickwidth         = 0pt,
    enlarge y limits  = 0.2,
    enlarge x limits  = 0.02,
    xmin = 29.3,
    xmax = 40,
    nodes near coords,
    nodes near coords style={/pgf/number format/.cd,fixed zerofill,precision=1},
    symbolic y coords = {EDA, 
                         Back-Translation,
                         {SwitchOut}, 
                         {Pervasive Dropout}, 
                         {Token Substitution}, },
    reverse legend,
    xtick = data,
    legend style={anchor=north east}
  ]
    \addplot+ coordinates { 
        (31.5,{Token Substitution}) 
        (32.0,{Pervasive Dropout})
        (31.5,{SwitchOut})
        (30.8,Back-Translation)  
        (32.6,EDA) 
    };
    \addlegendentry{Curriculum}
    \addplot+ coordinates { 
        (30.9,{Token Substitution}) 
        (31.2,{Pervasive Dropout})
        (31.2,{SwitchOut})
        (30.6,Back-Translation) 
        (32.1,EDA)  
    };
    \addlegendentry{Standard}
  \end{axis}
\end{tikzpicture}
\vspace{-4.6mm}
\caption{
Common text data augmentation techniques work better in the curriculum framework (two-stage) than standard single-stage training.
% Results shown are for triplet loss classification averaged over four datasets, in which 
A model with no data augmentation achieved a performance of 30.2\%.
}
\label{fig:different-techniques}
\end{figure}
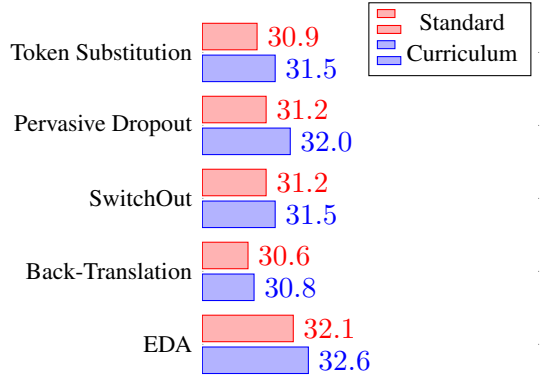
\subsection{For Various Augmentation Techniques}
\label{subsec:various}
As our experiments so far have focused on EDA augmentation \cite{wei-zou-2019-eda}, this section explores other common techniques in the curriculum framework:
\textbf{(1) Token Substitution}  replaces words with WordNet synonyms  \cite{NIPS2015_5782}; %\cite{miller1995wordnet};
\textbf{(2) Pervasive Dropout} applies word-level dropout with probability $p\shortequals0.1$ \cite{sennrich-etal-2016-edinburgh}; 
\textbf{(3) SwitchOut} replaces a token with a randomly token uniformly sampled from the vocabulary \cite{wang-etal-2018-switchout}; and 
\textbf{(4) Round-Trip Translation} translates text into another language and then back into the original language \cite{sennrich-etal-2016-improving}. %\footnote{We use the \small{\texttt{py-backtrans}} library.}
Figure \ref{fig:different-techniques} compares standard and two-stage curriculum results averaged over all datasets.
EDA improved performance the most, perhaps because it combines four token perturbation functions, creating more diverse noise compared with using a single operation.

\section{Related Work and Conclusions}
Our work combines curriculum learning, data augmentation, and triplet loss, and is inspired by prior work in these areas.
In vision, several papers have proposed reinforcement learning policies for data augmentation \cite{Cubuk_2019_CVPR,ho2019population}, and hard negative mining \cite{facenet,7780803} itself can be seen as a form of curriculum learning.
In NLP, the work of \citet{kumar-etal-2019-improving} is perhaps most similar to ours---they show that sampling strategies are key for improving performance with triplet loss networks.
We see our work as the first to explicitly analyze curriculum learning for data augmentation in text.

In closing, we have proposed a curriculum data augmentation framework that is simple yet provides empirical performance improvements, a compelling case for the combination of ideas explored.
Our approach exemplifies how data augmentation can create an artificial range of example difficulty that is helpful for curriculum learning, a direction that potentially warrants future research.

\section*{Acknowledgements}
We thank Ruibo Liu, Weicheng Ma, Jerry Wei, and Chunxiao Zhou for feedback on the manuscript.
We also thank Kai Zou for organizational support.

\newpage 
\bibliography{naacl2021}
\bibliographystyle{acl_natbib}

\clearpage
\appendix
\onecolumn

\begingroup
\begin{table*}[ht]
    \centering
    \small
    \begin{tabular}{l | c c c c }
        \toprule
        & HUFF & FEWREL & COV-C & AMZN \\
        & $c=41$ & $c=64$ & $c=87$ & $c=318$ \\
        \midrule
        \multicolumn{1}{l}{\textsc{Single-Stage Training}} \\
        Updates until convergence, no aug. (approx) & 4,000 & 8,000 & 4,000 & 15,000 \\
        Update until convergence, aug. (approx) & 10,000 & 10,000 & 8,000 & 20,000 \\
        Total updates & 15,000 & 15,000 & 15,000 & 25,000 \\
        \midrule
        \multicolumn{1}{l}{\textsc{Curriculum: Two-Stage}} \\
        Stage 1 updates & 4,000 & 6,000 & 4,000 & 8,000 \\
        Stage 2 updates & 11,000 & 9,000 & 11,000 & 17,000 \\
        Total updates & 15,000 & 15,000 & 15,000 & 25,000 \\
        \midrule
        \multicolumn{1}{l}{\textsc{Curriculum: Gradual}} \\
        % Schedule A: Stage 1 updates & 6,000 & 6,000 & 4,000 & 6,000 \\
        % Schedule A: Updates per stage in stages 2-6 & 4,000 & 4,000 & 3,000 & 6,000 \\
        % Schedule A: Total updates & 26,000 & 26,000 & 19,000 & 36,000  \\
        % \midrule
        Stage 1 updates & 6,000 & 6,000 & 6,000 & 10,000 \\
        Updates per stage in stages 2-6 & 6,000 & 6,000 & 4,000 & 8,000 \\
        Total updates & 36,000 & 36,000 & 26,000 & 50,000\\
        \bottomrule
    \end{tabular}
    \caption{
    Training schedules for single-stage training, two-stage curriculum training, and gradual curriculum training.
    }
    \label{tab:parameters}
\end{table*}
\endgroup
\section{Appendix}
\label{sec:appendix}
Table \ref{tab:parameters} shows the training schedules for single-stage, two-stage curriculum, and gradual curriculum training. 

All models in standard single-stage training (with and without augmentation) for the same dataset trained for the same number of updates; convergence typically took longer with augmentation compared to without augmentation.

Curriculum two-stage training employs a first stage of only original data and a second stage of augmented data, using the same number of updates as single-stage training in total.
We determined the number of updates in the first stage based on when training loss plateaued in the training plot for training with no augmentation.

The gradual curriculum starts with one stage of training with original data only and then increases the augmentation temperature by 0.1 in each of the following five stages.
To determine the number of updates in each stage, we examined training plots in preliminary experiments and increased the augmentation temperature (i.e., begun the next stage) whenever training loss plateaued.
Since our preliminary experiments already showed relatively strong performance improvements, we did not perform an extensive hyperparameter search or experiment with automatic scheduling, which could further improve performance.
As the gradual curriculum trains on more diverse set of augmented data, more updates are needed than in the single-stage and two-stage schedules.

For evaluation, we evaluate our models every 200 updates for COV-C and every 300 updates for HUFF, FEWREL, and AMZN, reporting the highest validation accuracy achieved during training.

In all models, we include 20\% original data whenever augmented data is used, in order to prevent catastrophic forgetting.

\end{document}